\documentclass[10pt,twocolumn,letterpaper]{article}
\usepackage{iccv} 
\usepackage{times}
\usepackage{epsfig}
\usepackage{graphicx}
\usepackage{amsmath}
\usepackage{amssymb}
\usepackage{graphicx}
\usepackage{balance}
\usepackage{xcolor}
\usepackage{makecell, multirow, booktabs}
\usepackage{enumitem}

\usepackage{listings}
\usepackage{wrapfig,lipsum,booktabs}
\setlist[itemize]{align=parleft,left=0pt..0.8em}

\usepackage[pagebackref=true,breaklinks=true,letterpaper=true,colorlinks,bookmarks=false]{hyperref}
\usepackage{authblk}
\usepackage{floatrow}
\usepackage[capitalize]{cleveref}
\crefname{section}{Sec.}{Secs.}
\Crefname{section}{Section}{Sections}
\Crefname{table}{Table}{Tables}
\crefname{table}{Tab.}{Tabs.}
\iccvfinalcopy
\ificcvfinal\fi

\newcommand\scalemath[2]{\scalebox{#1}{\mbox{\ensuremath{\displaystyle #2}}}}
\makeatletter
\renewcommand\AB@affilsepx{, \protect\Affilfont}
\makeatother

\begin{document}

	\newcommand{\sysname}{SeqNet}
	\newcommand{\algname}{SpaceEvo}
	\newcommand{\lz}[1]{{\textcolor{red}{\it Li: #1}}}	

\title{{\algname}: Hardware-Friendly Search Space  Design for Efficient INT8 Inference}

\author{ Li Lyna Zhang$^{1*\ddagger}$ Xudong Wang$^{2*\mathsection}$ Jiahang Xu$^1$ Quanlu Zhang$^1$ Yujing Wang$^3$ \newline Yuqing Yang$^1$ Ningxin Zheng$^1$ Ting Cao$^1$  Mao Yang$^1$\\\fontsize{10}{10} \selectfont{$^1$Microsoft Research, $^2$Shanghai Jiao Tong University, $^3$Microsoft}}

\maketitle
\def\thefootnote{*}\footnotetext{Equal contribution}
\def\thefootnote{$\ddagger$}\footnotetext{Corresponding author, email: lzhani@microsoft.com}
\def\thefootnote{$\mathsection$}\footnotetext{Work was done during the internship at MSRA}
\def\thefootnote{}\footnotetext{}
\ificcvfinal\thispagestyle{empty}\fi



\begin{abstract}
The combination of Neural Architecture Search (NAS) and quantization has proven successful in automatically designing low-FLOPs INT8 quantized neural networks (QNN). However, directly applying NAS to design accurate QNN models that achieve low latency on real-world devices leads to inferior performance. In this work, we find that the poor INT8 latency is due to the quantization-unfriendly issue: the operator and configuration (e.g., channel width) choices  in prior art search spaces lead to \textit{diverse} quantization efficiency and can slow down the INT8 inference speed. To address this challenge, we propose {\algname}, 
an automatic method for designing a dedicated, quantization-friendly search space for each target hardware. The key idea of {\algname} is to automatically search hardware-preferred operators and configurations to construct the search space, guided by a metric called \textit{Q-T score} to quantify how quantization-friendly a candidate search space is. We further train a quantized-for-all supernet over our discovered search space,  enabling the searched models to be directly deployed without extra retraining or quantization. Our discovered  models
establish new SOTA  INT8 quantized accuracy under various latency constraints, achieving up to 10.1\% accuracy improvement on ImageNet than prior art CNNs under the same  latency. Extensive experiments on diverse edge devices demonstrate that  {\algname} consistently outperforms existing manually-designed search spaces with up to 2.5$\times$ faster speed while achieving the same accuracy.

\end{abstract}
\vspace{-3ex}
\section{Introduction}
\vspace{-1ex}
INT8 Quantization\cite{datafreequant,8bitquant,lsq,lsqplus} is a widely used technique for deploying DNNs on edge devices by reducing 4$\times$ in model size and memory cost for full-precision (FP32) models. However, prior art DNN models achieve only marginal speedup from INT8 quantization (in Figure~\ref{fig:quant_model_benchmark}(a)), the still high latency after quantization making them difficult to deploy on latency-critical scenarios. Designing models that achieve high accuracy and low latency after quantization becomes the important but challenging problem.

Neural Architecture Search (NAS) is a powerful tool for automating efficient quantized model design~\cite{haq,dnas,spos,apq,mps}. 
Recently, OQAT~\cite{oqa} and BatchQuant~\cite{batchquant}  achieve remarkable search efficiency and accuracy by adopting a two-stage paradigm. The first stage trains a weight-shared quantized supernet assembling all candidate architectures in the search space. This allows all the sub-networks (subnets) to simultaneously reach comparable quantized accuracy as  when trained from scratch individually. The second stage uses typical search algorithms to find  subnets with best quantized accuracy under different FLOPs constraints.
This approach avoids the need to retrain each subnet for accuracy evaluation, greatly improving the search efficiency.

Though promising in optimizing FLOPs, we find that directly applying two-stage NAS to low quantized latency scenario leads to poor performance due to the  
\textit{quantization-unfriendly search space} issue: prior art search spaces cannot be well applied to quantization on diverse devices, as the current design  can unexpectedly \textit{hurt the INT8 latency}. 
Since INT8 quantization only offers a marginal speedup, we are forced to search for small-sized models to meet the latency requirements, which can unfortunately limit NAS to find better quantized models for edge devices. Then, a question naturally arise: \textit{Can we design a quantization-friendly search space, allowing NAS to  discover larger and superior models that meet the low INT8 latency requirements?}

We start by conducting an in-depth study  to understand 
the factors that determine INT8 quantized latency and how they affect search space design.  Our study shows: \textit{(1) both operator type and configurations (e.g., channel width) greatly impact the INT8 latency}; Improper selections can slow down the INT8 latency. For instance, Squeeze-and-Excitation (SE) \cite{se} and Hardswish~\cite{mobilenetv3} are widely-used operators  in  current  search spaces as it improves accuracy with little latency introduced. However, their INT8 inference speeds are \emph{slower} than FP32 inference on Intel CPU (Fig.~\ref{fig:convchannel}(a)), because the extra costs (\textit{e.g.}, data transformation between INT32 and INT8) introduced by quantization outweigh the latency reduction by INT8 computation.   \textit{(2) The quantization efficiency varies across different devices, and the preferred operator types can be contradictory.} 

The above study motivates us to design specialized quantization-friendly search spaces for each hardware, rather than relying on a single, large search space as seen in SPOS~\cite{spos} for all hardware, which provides different operator options per layer. This is because two-stage NAS requires the search space to adhere to a specific condition for training the supernet, wherein each layer must utilize a fixed operator. Our study indicates significant variations in optimal operators across different hardware. Thus, customizing the search space for each hardware is crucial for optimal results.
 However, designing such specialized quantization-friendly search spaces for various edge devices presents a significant challenge,  requiring expertise in both AI and hardware domains, as well as many trial and error attempts to optimize accuracy and INT8 latency for each hardware.

 In this paper, we propose {\algname}, a novel method for automatically designing specialized quantization-friendly search space for each hardware. The search space is comprised of hardware-preferred operators and configurations, enabling the search of larger and better models with low INT8 latency.  With the discovered search space, we  leverage two-stage quantization NAS to train a quantized-for-all supernet, and utilize evolution search~\cite{ofa} to find best quantized models under various INT8 latency constraints. Our approach  addresses three key challenges: (1) What is the definition of a quantization-friendly search space in terms of both quantized accuracy and latency? (2) How to automatically design a search space without human expertise? (3)  How to handle with the prohibitive cost caused by quality evaluation of a candidate search space?



 To address the first challenge, we  propose a latency-aware \textit{Q-T} score to quantify the effectiveness of a candidate search space, which measures the INT8 accuracy-latency quality of \textbf{top-tier} subnets in a search space. The behind intuition is that the goal of NAS is to search top subnets with better accuracy-latency tradeoffs.

Then, we introduce an evolutionary-based search algorithm that can effectively  search a quantization-friendly search space with highest \textit{Q-T} score. Searching a search space involves discovering \textit{a collection of model population} that contains billions of models, which is challenging and easily introduce complexity. To address this challenge, we propose to factorize and encode a search space into a sequence of elastic stages, which have flexible operator types and configurations. Through this design, the task of search space design is then simplied to find a search space with the optimal elastic stages, so that existing  search algorithms can be easily applied. Specifically, we design a stage-wise hyperspace to include many candidate search spaces and leverage aging evolution~\cite{real2019regularized} to perform random mutations of elastic stages for search space evolution. 
The  evolution is guided by maximizing the \textit{Q-T}  score. 

Finally, estimating the quality score (Q-T score) of a search space involves a costly training process for evaluating the accuracy of sub-networks, which presents a significant obstacle for our evolutionary algorithm. Naively adopting a two-stage NAS approach, training a supernet for each candidate search space~\cite{nse,autoformerv2}, is prohibitively expensive, taking  of thousands GPU hours. To address this issue, we draw inspiration from block-wise knowledge distillation~\cite{donna,dna} and propose a \textit{block-wise search space quantization scheme}. This scheme trains each elastic stage separately and rapidly estimates a model's quantized accuracy by summing block-level loss with a quantized accuracy lookup table, as shown in Fig.~\ref{fig:bkd}. This significantly reduces the training and evaluation costs, while providing effective accuracy rankings among search spaces. 
We summarize our contributions as follows:
\begin{itemize}
	\vspace{-1ex}
	\item We study the  INT8 quantization efficiency on real-world edge devices and find that the choices of operator types and configurations in a quantized model can significantly impact the INT8 latency, leaving a huge room for design optimization of quantized models.
	
	\item We propose {\algname} to automatically design a hardware-dedicated quantization-friendly search space and leverage two-stage quantization NAS to produce superior INT8 models under various latency constraints. 
	
	\item We present three innovative techniques that enable the first-ever efficient and cost-effective evolution search  to explore a search space comprising billions of models.
	\item Extensive experiments on two edge devices and ImageNet demonstrate that our automatically designed search spaces significantly surpass previous manually-designed search spaces. Our discovered models establish the new state-of-the-art INT8 quantized accuracy-latency tradeoffs. For instance, {\sysname}@cpu-A4, achieves 80.0\% accuracy on ImageNet, which is
	3.3ms faster with 1.8\% higher accuracy than FBNetV3-A. 
	Moreover, {\algname} produces superior tiny models, achieving up to 10.1\% accuracy improvement over the tiny ShuffleNetV2x0.5 (41M FLOPs, 4.3ms). 
	
	\end{itemize}

\vspace{-1.5ex}
\section{Related works}
\vspace{-1ex}
\noindent\textbf{Quantization} has been widely used for efficiency in deployment. Extensive efforts can be classified into post-training quantization (PTQ)~\cite{datafreequant,4bitptq} and quantization-aware training (QAT)~\cite{8bitquant,relaxedquant,lsq,lsqplus}. QAT generally outperforms PTQ in quantizing compact DNNs to  8bit and very low-bit (2, 3, 4bit) by finetuning the quantized weights. Despite their success, traditional quantization methods focus on minimizing accuracy loss for a given pre-trained model, but ignore the real-world inference efficiency. 

\noindent\textbf{NAS for Quantization}. 
Early works~\cite{haq,dnas,spos,apq,mps} formulates mixed-precision problem into NAS to search layer bit-width with a given architecture. Recently,  ~\cite{oqa,batchquant}  train a quantized-for-all supernet to search both architecture and bit-width. The searched models can be directly evaluated with comparable accuracy to train-from-scratch. However, little attention is paid on optimizing quantized  latency on real-world devices. Through searching quantization-friendly search space, our discovered quantized models can achieve both high accuracy and low latency.


\noindent\textbf{Search Space Design}. 
 Starting from ~\cite{proxylessnas}, the manually-designed MBConv-based space becomes the dominant in most NAS works~\cite{proxylessnas,ofa,bignas,attentivenas}.
 RegNet~\cite{regnet} is the first to present  standard guidelines to optimize a search space by each dimension.  Recently, ~\cite{angle-based,asap,pcnas,padnas,nse,autoformerv2} propose to  shrink to a better compact search space by either pruning unimportant operators or configurations.  
  However, these works focus on optimizing the accuracy and little attention is paid on quantization-friendly search space design. Our work is the first lightweight solution towards this direction.

\section{On-device Quantization Efficiency Analysis}
\label{sec:analysis}
\vspace{-1ex}



To understand what factors  lead to quantization-unfriendly issue, we conduct a comprehensive study on two widely-used edge devices equipped with high-performance inference engine: an Intel CPU device supported with VNNI instructions and onnxruntime~\cite{onnxruntime} (abbr Intel CPU) and a Pixel 4 phone CPU with TFLite 2.7~\cite{tflite} (abbr Pixel 4). Note that we follow existing practices~\cite{nnmeter,spacestudy,vitstudy} to measure the latency. 
 We reveal key observations as follows:
 
\begin{figure}[t]
	\centering
	\includegraphics[width=1\columnwidth]{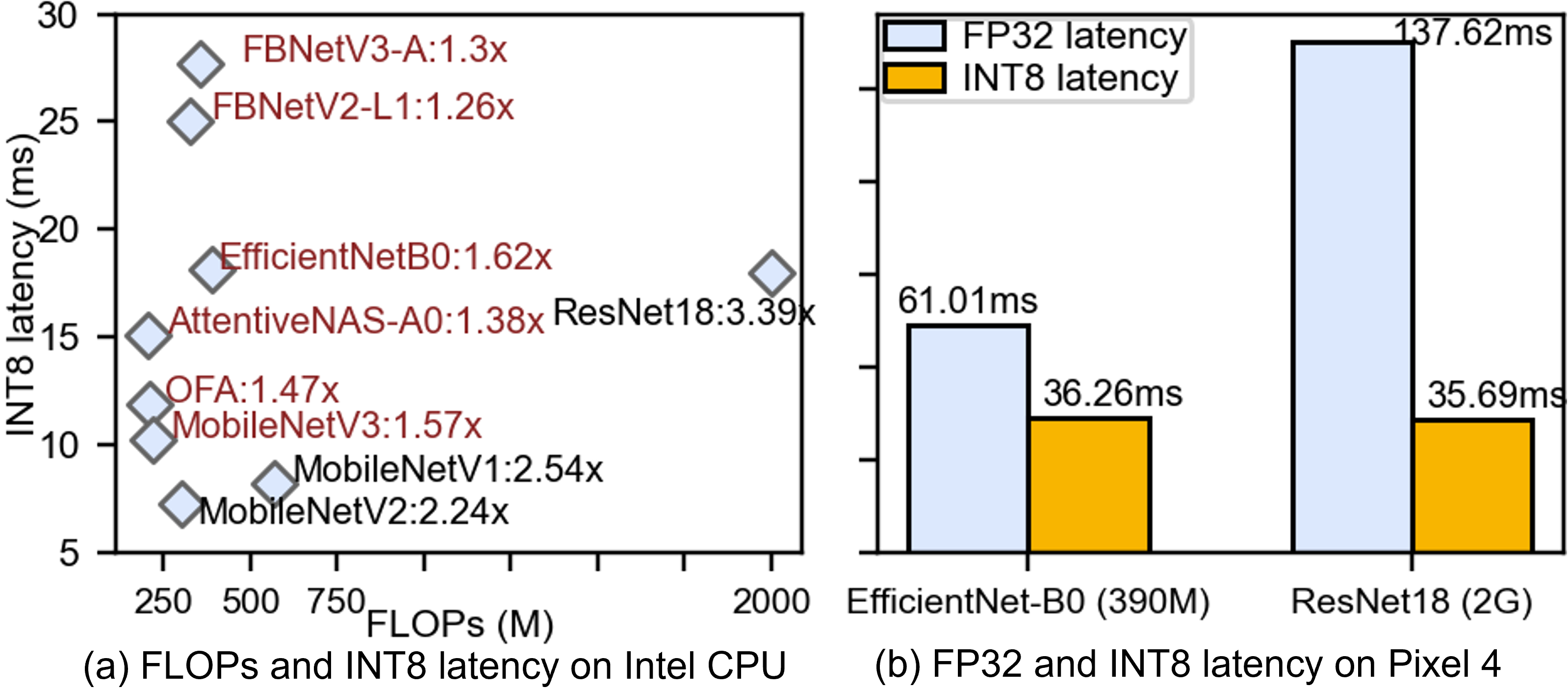}
	\vspace{-4ex}
	\caption{INT8 latency and speedups (annotated) for SOTA models. FLOPs and FP32 latency are not good indicators of INT8 latency; Compact models have very limited INT8 speedup ($\sim$1.5$\times$). }
	\label{fig:quant_model_benchmark}
\end{figure}

\vspace{2px}
\noindent\textbf{Observation 1}: \textit{FP32 latency and FLOPs are not good indicators of INT8 latency. }

 To deploy  on edge devices, a common belief is that a compact model with low FLOPs or FP32 latency is preferred than a larger model. However,  Fig.~\ref{fig:quant_model_benchmark} shows that neither of them is a good indicator of INT8 latency. In Fig.~\ref{fig:quant_model_benchmark}(b), a very large model (ResNet18) can be even faster than a compact model (EfficientNetB0) after quantization. Moreover, the recent SOTA compact models searched by  OFA~\cite{ofa} and AttentiveNAS~\cite{attentivenas} all have marginal INT8 speedups, suggesting that optimizing FLOPs and FP32 latency can not lead to lower INT8 latency.  

\noindent\textbf{Observation 2}:  \textit{The choices of operators' types and configurations greatly impact the INT8 latency. } 

\begin{figure}[t]
\centering
\includegraphics[width=1\columnwidth]{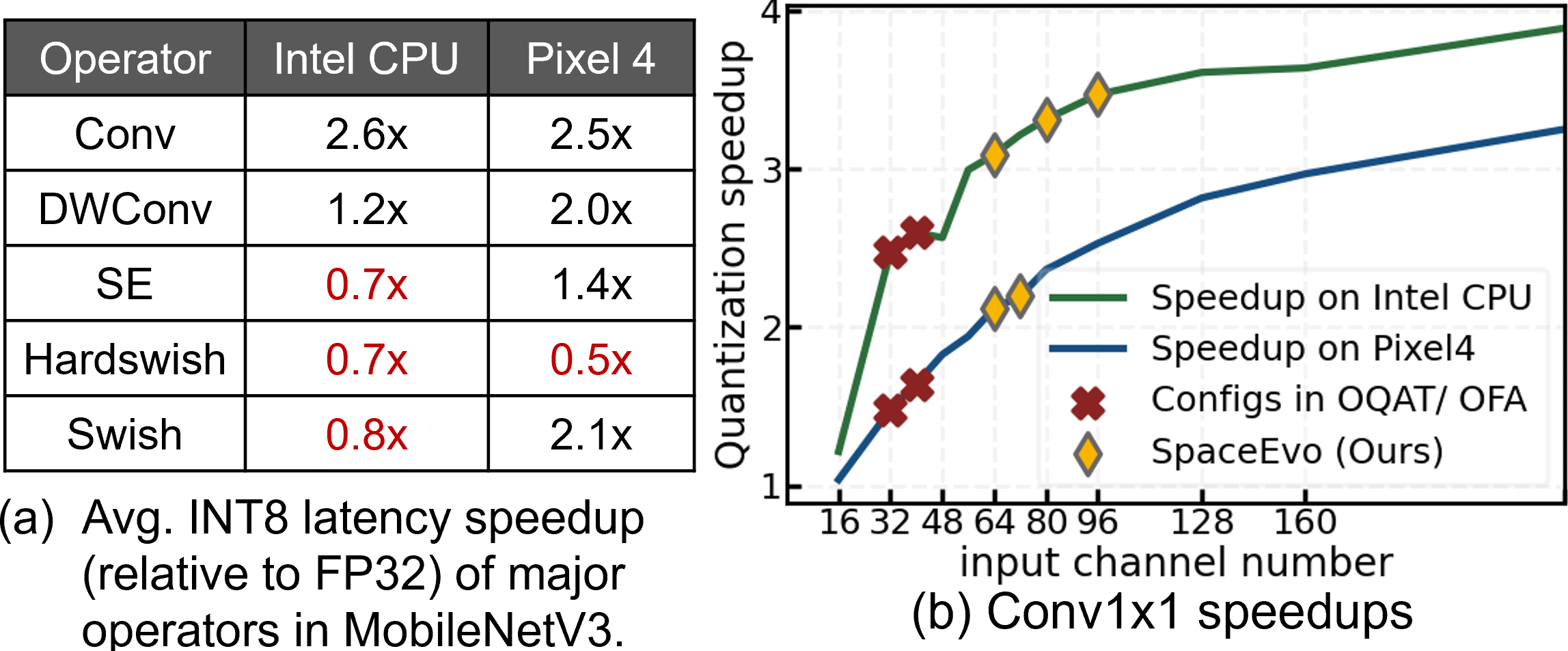}
\vspace{-3ex}
\caption{ (a) The choice of operator type leads to significantly different quantized speedup. (b) Conv1x1 speedups with various channel numbers. Config: $HW$=28, $C_{out}$=4x$C_{in}$(expand=4). }

\label{fig:convchannel}
\end{figure}
The prior art search spaces adopted in recent two-stage NAS works 
are MobileNetV2 or MobileNetV3 based chain-structures, where each search space comprises a sequence of blocks (stages). The block type is \textit{fixed} to the  MBConv 
 and is allowed to search from a \textit{handcraft range of hyper-parameter configurations} including kernel size, expansion ratio, channel width and depth, which  are designed with human wisdom. For instance, many works~\cite{proxylessnas,attentivenas} observe that edge-regime CNNs prefer deeper depths and narrower channels, and manually set  small channel numbers but large depths in the search space.  


However, we find that many block type and configuration choices in current search spaces unexpectedly slow down the INT8 latency. We first study the operator type impact in Fig.~\ref{fig:convchannel}(a). SE and Hardswish are lightweight operators in edge-regime search spaces, but their INT8 inference becomes slower on Intel CPU. Compared to Conv, DWConv can greatly reduce the FLOPs, but it benefits less from INT8 quantization.  The root cause is that quantization introduces extra cost, such as (1) data transformation between INT32 and INT8~\cite{spacestudy}, and (2) additional computation caused by scaling fators and zero points~\cite{jacob2018quantization}. If the operator has low data-reuse-rate, such as the activation functions (Hswish), the extra cost may outweigh the latency reduction by the low-bit computation. For high data-reuse operators (Conv), this cost is amortized and thus achieve large speedup~\cite{spacestudy}. 


Besides the operator type, the configuration choices also determine the quantization efficiency. Fig.~\ref{fig:convchannel}(b) shows the  speedups of Conv1$\times$1 under various channel widths. Results show that small channel widths in OFA search space cannot benefit well from quantization.  This is because the additional quantization cost  has a large impact when the channel width is small, limiting the latency acceleration.  In contrast,  {\algname} can automatically design a search space with larger channel widths for better efficiency.



\noindent\textbf{Observation 3}: \textit{Quantization-friendly settings are diverse and contradictory across devices.} 

In Fig.~\ref{fig:convchannel}, we also observe that the quantization-friendly operators are different and can be contradictory on diverse devices. For instance,  Swish achieves a 2$\times$ speedup on Pixel4, but it is a quantization-unfriendly operator on CPU with a 0.8$\times$ slowdown.  
The reason is that quantization speedups are highly dependent on the inference engines and hardware~\cite{spacestudy,9638444}.  Intel VNNI supports the VPDPBWSD hardware instruction~\cite{vnni}, which fuses three instructions into one to speedup INT8 computation. Without VNNI, INT8 hardly gains speedup on Intel CPUs. 
Moreover, the implementations in inference engines have to fully utilize hardware instructions for latency reduction.
For example, onnxruntime does not implement a quantization kernel for Hardswish. Even on a VNNI CPU, the use of Hardswish in a quantized model slows down the  latency.

\vspace{2pt}
\noindent\textbf{The need for hardware specialized search space}.
The above observations suggest that there is no single structure (block types in a model) that is optimal for quantization on all hardware. This poses a challenge for the two-stage NAS paradigm, as the supernet training requires all models in the search space to share an isomorphic structure.  To address this issue, 
our work proposes to design a specialized quantization-friendly search space for each hardware. Each search space is tailored to the unique characteristics of the hardware and includes an optimal structure with elastic depths, widths, and kernel sizes. 

\section{Methodology}
\begin{figure*}[t]
	\centering
	\includegraphics[width=1.05\textwidth]{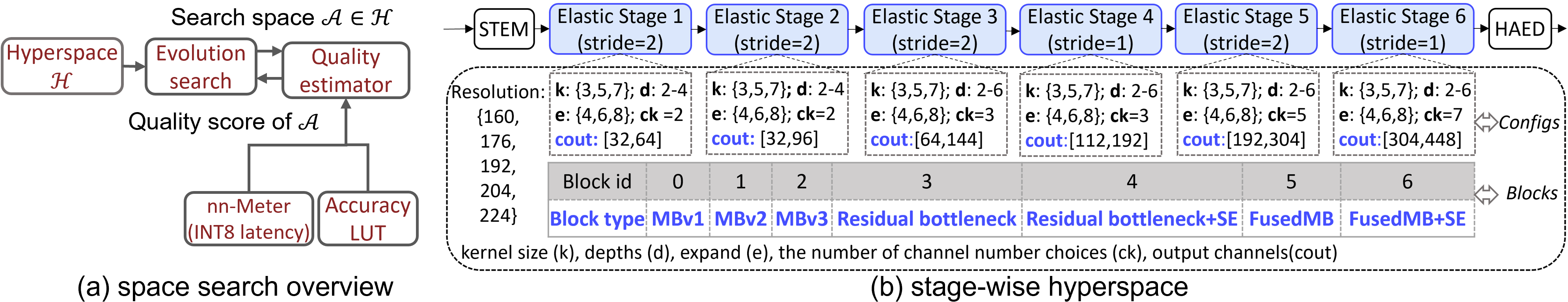}	
	\vspace{-4.5ex}
	\caption{(a) We simplify space search into model search process; (b) Illustration of  our hyperspace. A sampled search space is encoded by a sequential elastic stages.  Contents in blue are searched: an elastic stage can search its block type and channel number list. }
	\label{fig:hyperspace}
\end{figure*}
\vspace{-1ex}
\subsection{The Core Design Concept}
\vspace{-1ex}
 In this section, we present our methodology for automatically designing a specialized quantization-friendly search space for any target hardware.  Different from architecture search, where the goal is to find the single best model from the space,  we aim to discover a \textit{model population} that contains billions of accurate and INT8 latency-friendly architectures. 
We draw inspiration from the neural architecture search process and propose to use an evolutionary search algorithm to explore such a quantization-friendly model population. To achieve this, we introduce {\algname}, which is built on the following three techniques.

First, we need a metric that quantifies how quantization-friendly a candidate search space is. We define a Q-T score that is efficiently measured by top-tier subnets' INT8 accuracy-latency (\cref{sec:qt}).

Second, existing evolutionary search algorithms are designed for searching a single model architecture rather than a large search space encompassing billions of architectures. we propose a novel approach that we call the "elastic stage." By factorizing the search space into a sequence of elastic stages ((\cref{sec:elasticstage}), we enable traditional aging evolution methods, such as the aging evolution~\cite{real2019regularized}, to be directly applied to search the space (\cref{sec:search}). 

Third, searching a search space with a maximum Q-T score can be prohibitively costly since the corresponding supernet must be trained from scratch for accuracy evaluation. We propose a block-wise search space quantization scheme to significantly reduce the training cost(\cref{sec:bkd}).

\subsection{Search Space Quality Score}
\label{sec:qt}
\noindent\textbf{Latency-aware space quality of Q-T score}. Before space search, we need a score to quantify how quantization-friendly a search space is, which serves as the search objective. Since our ultimate goal is to search the best quantized models from the searched space, we treat a space with good quality if its \textbf{top-tier} subnets achieve optimal quantized accuracy under latency constraints $T$. Due to the fact that real-world applications usually have different deployment requirements, we use \textbf{multiple INT8 latency constraints} to measure a space's quality. 
For search space $\mathcal{A}$ and  a set of quantized latency constraints $T_{1,...,n}$,  we treat every constraint equally important, and define Q-T score as   the  sum of each constraint:  $\mathcal{Q}(\mathcal{A}, T_{1,...,n})$= $\mathcal{Q}(\mathcal{A}, T_{1})$+$\mathcal{Q}(\mathcal{A}, T_{2})$+..., $\mathcal{Q}(\mathcal{A}, T_{n})$,  $\mathcal{Q}(\mathcal{A}, T_{i})$ is defined as:
\begin{equation}
	\label{eq:score}
	\begin{aligned}
		\displaystyle \mathcal{Q}(\mathcal{A}, T_{i})= \mathbb{E}_{\alpha\in \mathcal{A}, LAT(\alpha)\leq T_i}[ Acc_{int8}(\alpha) ]
	\end{aligned}
\end{equation}
where $\alpha$ denotes a top-tier (best searched) subnet in $\mathcal{A}$ and
$Acc_{int8}(\alpha)$ is its  top-1 quantized accuracy evaluated on ImageNet validation set, $LAT(\alpha)$ predicts the quantized latency on target device.

However, it's non-trivial to obtain the top-tier subnets from a candidate search, as if often involves an expensive full architecture search process.  We adopt a \textit{zero-cost} policy.  Specifically, we randomly sample 5k subnets and select top 20   that under the latency constraints as the top-tier models to approximate the expectation term. The top 20 subnets are rapidly identified through the use of an accuracy look-up-table and a quantized latency predictor (\cref{sec:bkd}).

\subsection{Elastic Stage and Problem Formulation}
\vspace{-1ex}
\label{sec:elasticstage}
 We  observe that existing two-stage NAS adopt a chain-structured search space~\cite{ofa,bignas,attentivenas}, which  can be factorized as a sequence of STEM, HEAD and $N$ searchable stages. Each stage defines  a range of configurations $c$ (e.g., kernel size, channel width, depth) for a specific block type $b$, and allows NAS to find the optimal architecture settings. 

\noindent\textbf{Elastic stage}. Without loss of generality, we define a stage structure in a search space  as elastic stage $E_{b,c}$, which has elastic configurations $c$ for a fixed block $b$. 
Suppose a search space $\mathcal{A}$ has $N$ stages, it can be modularized as:
\begin{equation}
		\begin{aligned}
			\label{eq:searchspace}
\displaystyle \mathcal{A}=STEM \circ E_{b,c}^{1},... \circ E_{b,c}^{N} \circ HEAD
\end{aligned}
\end{equation}
For instance, the search space in OQAT~\cite{oqa} and BatchQuant~\cite{batchquant} can be factorized as 6 elastic stages and STEM (first Conv) and a classification head. Each elastic stage represents a set of configuration choices for the MBv3 block. Through the definition of elastic stage, we can simply use ~\cref{eq:searchspace} to denote the contents of a  model popuation.  

\vspace{2pt}
\noindent\textbf{Problem definition}. Operator type $b$ and  configuration $c$ are two crucial objectives when searching quantization-friendly search space.
Through the definition of elastic stage, the task of space search  can be simplified to find a search space with optimal elastic stages, which has a similar goal with  neural architecture search.
We formulate our problem as:

\begin{equation}
	\label{eq:problem}
	\begin{aligned}
		\scalemath{0.86}{\mathcal{A}(E_{b,c}^{1}\circ E_{b,c}^{2}\circ...\circ E_{b,c}^{N})^*=\mathop{\arg\max}\limits_{E_{b,c}^i\in \mathcal{H}^i}{\mathcal{Q}(\mathcal{A}(E_{b,c}^{1}\circ E_{b,c}^{2}\circ ...\circ E_{b,c}^{N}), T)}
	}\end{aligned} 
\end{equation}
where $\mathcal{A}(\cdot)$  denotes the search space, and $E_{b,c}^i$ is the $i^{th}$ elastic stage of  $\mathcal{A}(\cdot)$. $\mathcal{H}$ denotes the hyperspace  which covers all possible search spaces.  $\mathcal{Q}$ is the Q-T score for measuring a  search space's quality. Given the constraints $T$ (i.e., a set of targeted quantized latency),
{\algname} aims to find the optimal  elastic stages ($E_{b,c}^{1}\circ E_{b,c}^{2}\circ...\circ E_{b,c}^{N}$)$^*$ from the $1^{st}$ to $N^{th}$ stage for $\mathcal{A}^*$ that has the maximum Q-T score: the top-tier quantized  models  can achieve best accuracy under the constraint $T$. Fig.~\ref{fig:hyperspace}(a) illustrates the overall process.  In this work, we focus on the latency of INT8 quantized models.  Our approach can be generalized to lower bits once they are supported on standard devices.

\subsection{Searching the Search Space}
\label{sec:search}
We now describe our evolutionary search algorithm that solves the problem in \cref{eq:problem}.

\noindent\textbf{Hyperspace design}. Analogous to  NAS, hyperspace $\mathcal{H}$  defines which search space a search algorithm might discover. Defining a hyperspace to cover many candidate search spaces for space search is a second-order problem for NAS, which can easily introduce high complexity. Fortunately, we can easily construct a large hyperspace through search space modularization in ~\cref{eq:searchspace}.

We construct a large hyperspace in Fig.~\ref{fig:hyperspace}, in which 
a search space can be encoded by $N$=6 sequential elastic stages along with STEM and HEAD.
We search the following two dimensions for an elastic stage:
\vspace{-1ex}
\begin{itemize}
	\item Block (operator) type $b$:  MBv1~\cite{mobilenet}, MBv2~\cite{mobilenetv2}, MBv3~\cite{mobilenetv3}, residual bottleneck~\cite{resnet}, residual bottleneck with SE, FusedMB~\cite{efficientnetv2} and FusedMB with SE. Conv is the major operator in residual bottleneck and FusedMB, thus they are quantization-friendly blocks; the efficiency of MB blocks relies on the device. For example, DWConv and SE are less quantization-efficient on Intel CPU. 
	\item Output channel width list $cout$. In Sec.~\ref{sec:analysis}, we observe that quantized models can better utilize hardware  under a larger channel number setting. However, directly increasing the channel numbers will also lead to longer latency. Therefore,  we search the optimal stage-wise channel width list $cout$: $\{w_{min}^*, ..., w_{max}^*\}$, which provides better channel width choices for final INT8 model search. 
	Specifically, as described in Fig.~\ref{fig:hyperspace}(b), we  predefined a wide range of $[w_{min}, w_{max}]$ by enlarging the channel widths in existing search spaces, and allow each elastic stage to choose a subset of $cout$ from $[w_{min}, w_{max}]$.  
	
\end{itemize}
\vspace{-1ex}
Besides channel widths, other configuration dimensions (e.g., kernel size) also impact a model's quantized latency. However, searching all dimensions leads to a large amount of   choices in one stage, which exponentially enlarges the hyperspace size. Fortunately, other dimensions usually have a small space (e.g., kernel size selects from \{3,5,7\}). It's easy to find the optimal  value for a model in the final NAS process. Therefore, we follow existing practices to configure the choices of kernel size, depth, and expand ratios.  Our final searched INT8 model architectures {\sysname} suggest that the optimal quantization-friendly kernel sizes and expand ratios are chosen. For example, 
kernel size of 3×3 brings more INT8 latency speedups for DWConv, and it is the dominate kernel size choice in DWConv related blocks. 

Suppose that a stage has $m$ choices of channel widths, there would be 7 (operator types) $\times m$ candidates for each stage. In total, for a typical search space with $N=6$ stages, the hyperspace has $\sim$10$^9$ candidate search spaces, which is extremely large and poses challenges for efficient search.

\vspace{2pt}
\noindent\textbf{Evolutionary space search}.  The structure of hyperspace is similar to a typical model search space in NAS~\cite{spos}, so we can easily apply existing NAS search algorithms.  Taking this advantage, we leverage  aging evolution~\cite{real2019regularized} to search the large hyperspace. We first randomly initialize a population of $P$ search spaces, where each sampled space is encoded as  ($E_{b,c}^{1}\circ E_{b,c}^{2}\circ...\circ E_{b,c}^{N}$). Each individual  is rapidly evaluated with  Q-T  score. After this, evolution improves the initial population in mutation iterations. At each iteration, we sample $S$ random candidates from the population and select the one with highest score as the \textit{parent}. Then we alternately mutate the \textit{parent} for block type and widths to generate two \textit{children} search spaces. For instance, suppose the $i^{th}$  stage $E_{b,c}^{i}$   is selected for mutation, we first randomly modify its block type and produce $E_{b^{*},c}^{i}$ for child 1, then we mutate the widths and produce   $E_{b,c^{*}}^{i}$ for child 2. We evaluate their Q-T scores and add them to current population. The oldest two are removed for next iteration.
After all iterations finish, we collect all the sampled space and select the one with best score as the final search space.

	\vspace{-1ex}
	\subsection{Efficient Search Space Quality Evaluation}
	\vspace{-1ex}
	\label{sec:bkd}
	\begin{figure}[t]
		\centering
		\includegraphics[width=1.01\columnwidth]{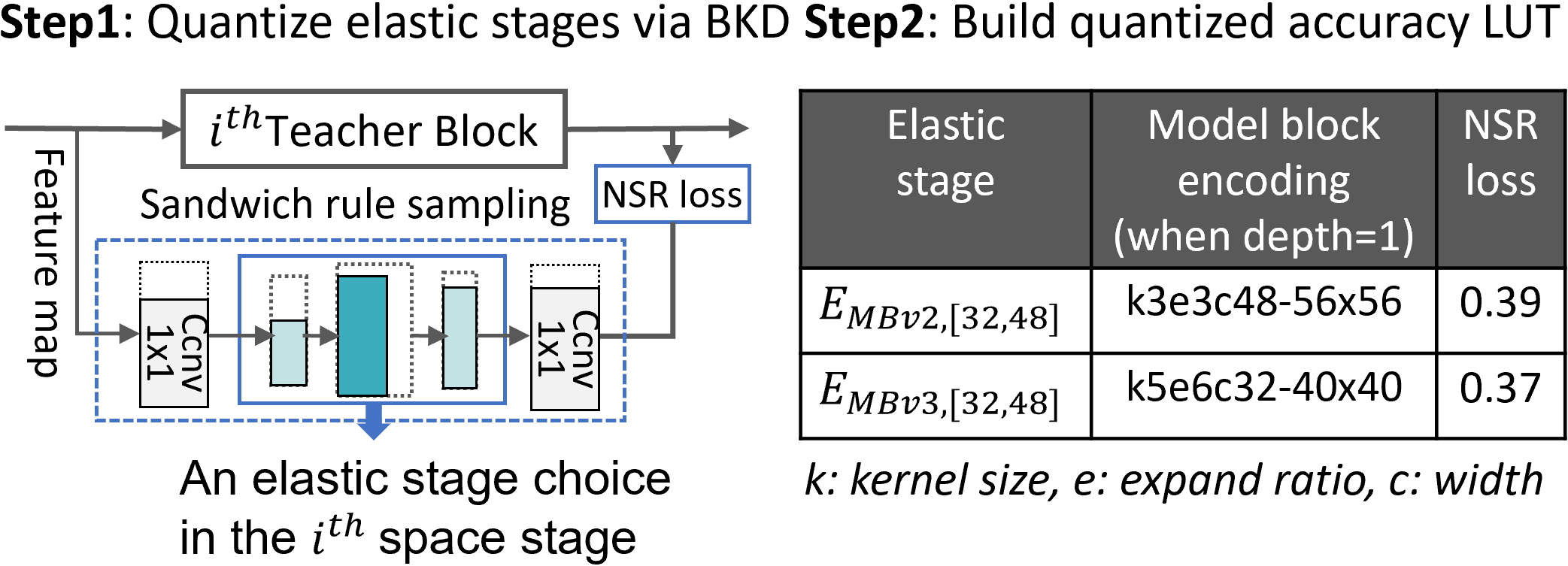}	
		\vspace{-4ex}
		\caption{ We adopt block-wise knowledge distillation to  reduce search space quality evaluation cost. We add two linear transformation layers (Conv 1x1) to match teacher's feature map widths. }
		\label{fig:bkd}
	\end{figure}
	
	We now address the  efficiency challenge caused by Q-T score evaluation. The most accurate  evaluation  is to get accuracy by training a supernet (search space) from scratch and measure latency on target device. However, it's impractical to conduct large-scale search due to the prohibitive cost. For example, it costs more than 10 days to train a supernet on 8 V100 GPUs~\cite{bignas}. To reduce the cost, we  build an accurate INT8 latency predictor by nn-Meter~\cite{nnmeter}, then we introduce block-wise quantization scheme.

	\noindent\textbf{Block-wise knowledge distillation (BKD)} is firstly proposed in DNA~\cite{dna} and then further improved in DONNA~\cite{donna}. It originally uses block-wise representation of existing models (teacher) to supervise a corresponding student model block. In our work, we extend BKD to
	supervise the training of all elastic stages (each contains a large amount of blocks). Since a stage size is much smaller than search space size,  the training time is greatly reduced. 

	
	Fig.~\ref{fig:bkd} illustrates the BKD process. In the first step, 
	we  use EfficientNet-B5 as the teacher, and separately train each elastic stage to mimic the behavior of corresponding teacher block by minimizing the NSR loss~\cite{donna} between their output feature maps. Specifically, the $i^{th}$ stage receives the output of $(i-1)^{th}$ teacher block as the input and is optimized to predict the output of  $i^{th}$ teacher block with NSR loss.
	Since an elastic stage contains many blocks with different channel widths, we add two learnable linear transformation layers at the input and output for each elastic stage to match teacher's feature map shape. Moreover, we adopt sandwich rule ~\cite{bignas} to sample four paths to improve the training efficiency. Each elastic stage is firstly trained for 5 epochs and then performed 1 epoch LSQ+~\cite{lsqplus} for INT8 quantization. 
	
	
	
\noindent\textbf{Accuracy lookup table}.	In the second step, we construct a INT8 accuracy lookup table to reduce the evaluation cost.  Specifically, we evaluate all possible blocks in each elastic stage and record their NSR losses on the validation set in the lookup table. The quantized loss of a model is estimated by summing up the NSR loss of all its blocks by rapidly looking up each elastic stages from the table. We inverse the measured loss to approximate the quantized accuracy for Q-T score evaluation.
	
	In our work, the BKD and lookup table construction can be sped up in a parallel way and finished in 1 day, which amounts a one-time cost before aging evolution search.

\section{Evaluation}

\noindent\textbf{Setup}. We evaluate our method on ImageNet-1k dataset~\cite{imagenet} and two popular edge devices. The INT8 latency constraints  are \{8, 10, 15, 20, 25\} \textit{ms} for Intel CPU, and \{15, 20, 25, 30, 35\} \textit{ms} for Pixel4.  For each device, we search 5k search spaces in total and return the one with highest Q-T score. The population size $P$ is 500 and sample size $S$ is 125. 

Once {\algname} discovers a quantization-friendly search space for the target device, we train a \textit{quantized-for-all} supernet. 
We start by pretraining a full-precision supernet without quantizers on ImageNet for 360 epochs. We adopt the sandwich rule and inplace distillation  in BigNAS~\cite{bignas}.  Then, we perform quantization-aware training (QAT) on the trained supernet for 50 epochs, which follows the same training protocol (i.e., sandwich rule and inplace distillation). We use LSQ+ as the QAT algorithm for better quantized accuracy.
To derive INT8 model for deployment, we use the evolutionary search in OFA~\cite{ofa} to search 5k models for various given INT8 latency constraints. We list out detailed training settings in supplementary materials.  In the following, we refer to the two searched spaces as  \textit{{\algname}@CPU} and \textit{{\algname}@Pixel4}, the searched model families are  \textit{{\sysname}@cpu} and \textit{{\sysname}@pixel4}.

 
 
 
 
\vspace{-1ex}
\subsection{The Effectiveness of {\algname}}

\begin{figure}[t]
	\centering
	\includegraphics[width=1\linewidth]{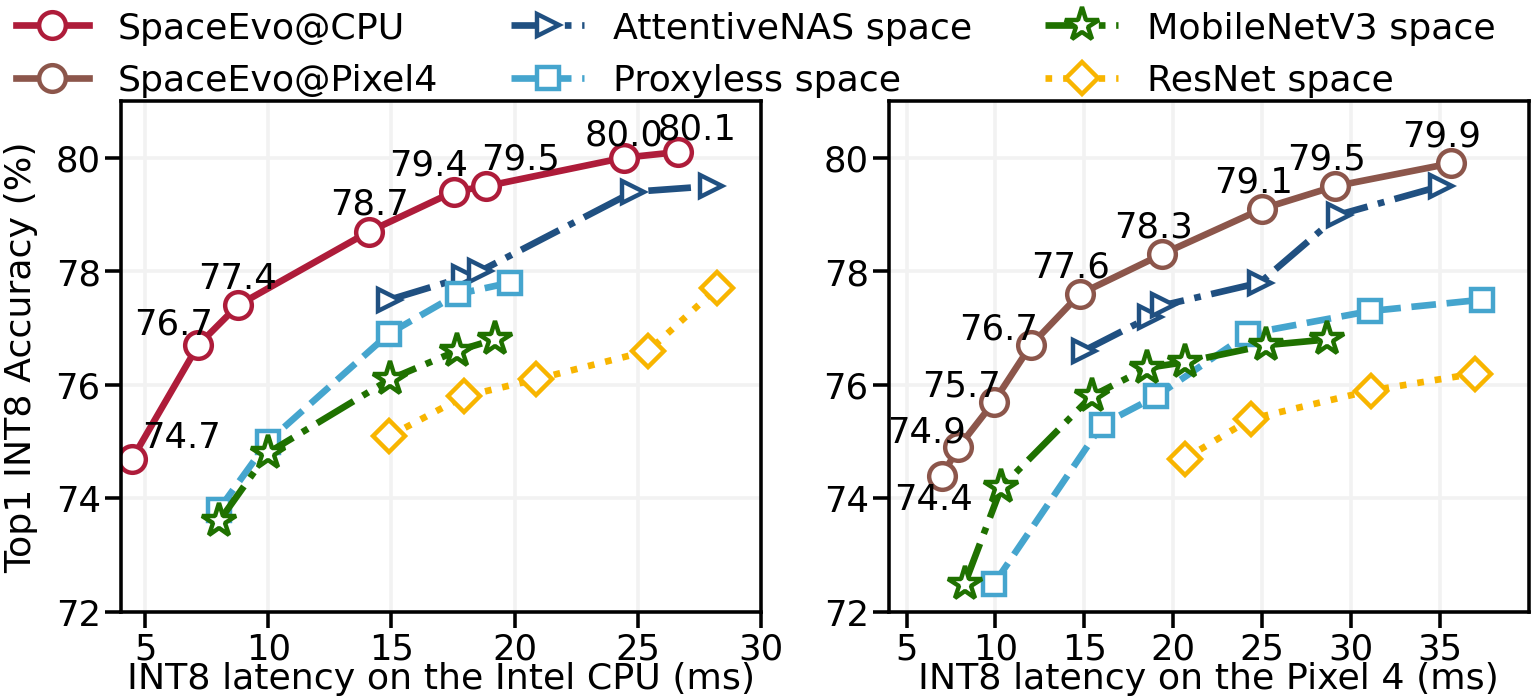}	
	\vspace{-4ex}
	\caption{Best searched INT8 models with comparison to state-of-the-art NAS search spaces. Our searched spaces are proven to be the most quantization-friendly for the target device.}
	\label{fig:searchspace_comparison}
\end{figure}

\noindent\textbf{Comparison with SOTA search spaces}. To demonstrate the high-performance of our searched spaces,  we compare  with prior art manually-designed search spaces including: (1) MobileNetV3 search space that is adopted in two-stage quantization NAS  OQAT~\cite{oqa} and BatchQuant~\cite{batchquant}; (2)  ProxylessNAS and AttentiveNAS search spaces that achieve superior performance on mobile devices; and (3) ResNet50 search space proposed by OFA that is a  handcraft quantization-friendly space on our two devices. 
For fair comparison, we use one supernet training and QAT receipt for all search spaces.    We conduct evolutionary search to compare the  best-searched models from each search space. 
We use the random seed of 0.  For all experiments, search space is the only difference.

 Fig.~\ref{fig:searchspace_comparison} compares the best searched INT8 models from different search spaces. {\algname}@CPU and {\algname}@Pixel4 consistently deliver superior quantized models than state-of-the-art search spaces. Under the same latency, the best quantized models from {\algname}@cpu significantly surpass the existing state-of-the-art search spaces with +0.7\% to +3.8\% (+0.4\% to +3.2\% on Pixel4 ) higher accuracy. Moreover, our search space is the only one that is able to deliver superior quantized models under both extremely low (only$\sim$5ms) and large latency constraints.   


\begin{figure}[t]
	\centering
	\includegraphics[width=0.8\columnwidth]{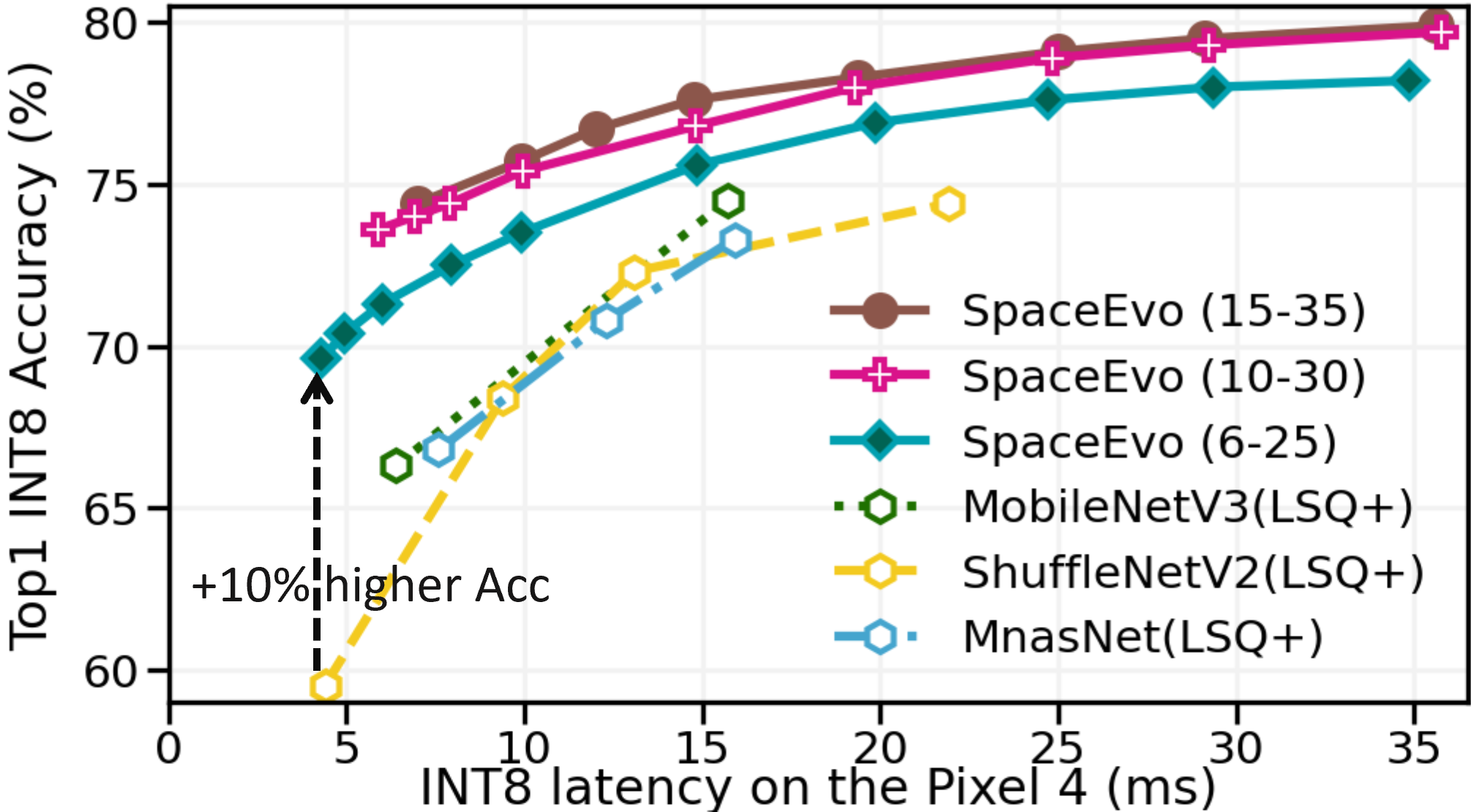}
	\vspace{-1ex}
	\caption{Search space design under diverse INT8 latency constraints. {\algname} (6-25 ms) delivers superior tiny INT8 models.}
	\label{fig:smallspace}
\end{figure}
\begin{figure}[t]
	\centering
	\includegraphics[width=1\columnwidth]{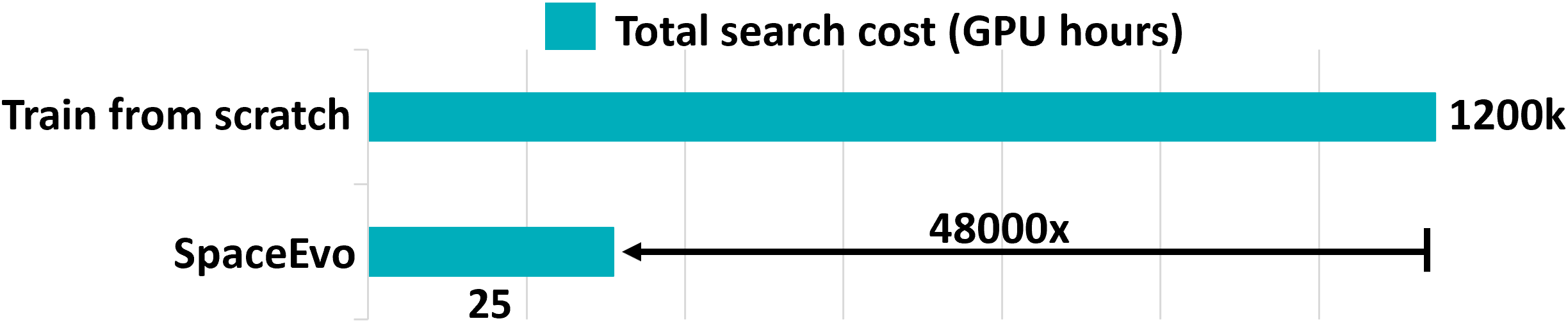}
	\vspace{-4ex}
	\caption{Search cost measured on 8 Nvidia V100 GPUs. }
	\label{fig:searchcost}
\end{figure}

\noindent\textbf{{\algname} under diverse latency constraints.} We extensively study the effectiveness of {\algname} under different latency constraints. Specifically, we perform  space search under two tight constraints of \{10, 15, 20, 25, 30\}ms and \{6, 10, 15, 20, 25\}ms on Pixel4. The results are shown in 
Fig.~\ref{fig:smallspace}. Our proposed method can handle the diverse latency requirements and produce  high-quality spaces.  As expected, the searched spaces under 10-30ms and 6-25ms have much more low-latency quantized models.

 To further verify the effectiveness of these low-latency models, we compare with existing SOTA tiny models. Significantly, even under the extremely low latency constraints of 6-25 ms, our searched space delivers very competitive tiny quantized models. Compared to the smallest model ShuffleNetV2x0.5, we can achieve +10.1\% higher accuracy under the same latency of 4.3 ms.


\begin{table}
		\centering
		\fontsize{8.2}{8.2} \selectfont
		\caption{ImageNet results compared with SOTA  quantized models on two devices. $^*$: latency compared to FP32 inference.  }
		\vspace{-2.5ex}	
		\begin{tabular}	
			{@{\hskip1pt}c|c|cc|c@{\hskip1pt}c@{\hskip1pt}}			
			\toprule
			\multicolumn{6}{c}{\textbf{(a) Results on the Intel VNNI CPU with onnxruntime}}\\
			\multirow{2}{*}{Model}  &Acc\%& \multicolumn{2}{c|}{CPU Latency}&Acc\%& \multirow{2}{*}{FLOPs} \\
			&\textbf{ INT8}& \textbf{INT8}&\textbf{speedup$^*$} &   FP32&\\ 
			
			\midrule
			MobileNetV3Small & 66.3&4.4 ms &1.1$\times$  & 67.4 &56M \\
			\textbf{	\sysname @cpu-A0}   & \textbf{74.7} & \textbf{4.4 ms} & \textbf{2.0$\times$} &74.8& 163M\\
			\hline
			MobileNetV2 &71.4 &7.3 ms &2.2$\times$ &72.0&300M \\
			
			ProxylessNAS-R &74.6  &8.8 ms  &1.8$\times$&74.6 &320M \\
			OQAT-8bit & 74.8 & 9.8 ms &1.8$\times$ &75.2&214M \\
			MobileNetV3Large  &74.5&10.3 ms &1.5$\times$&75.2 &219M  \\
			
			OFA (\#25)  &75.6&11.2 ms & 1.5$\times$ &76.4&230M\\
			\textbf{	\sysname @cpu-A1}    & \textbf{77.4} & \textbf{8.8 ms} & \textbf{2.4$\times$}&77.5 & 358M\\
			\hline
			APQ-8bit& 73.6&15.0 ms & 1.5$\times$& 73.6& 297M\\
			
			AttentiveNAS-A0  &76.1 &15.1 ms &1.4$\times$& 77.3 & 203M\\
			OQAT-8bit & 76.3 & 14.9 ms &1.7$\times$ &76.7& 316M\\
			EfficientNet-B0 &76.7&18.1 ms & 1.6$\times$&77.3 &390M\\
			\textbf{	\sysname @cpu-A2} &\textbf{78.5} & \textbf{14.1 ms} & \textbf{2.4$\times$}  &78.8 & 638M\\
			\hline
			APQ-8bit  & 74.9&20.0 ms & 1.5$\times$&75.0&393M\\
			OQAT-8bit  & 76.9 & 19.5 ms &1.6$\times$&77.3&405M \\
			AttentiveNAS-A1&77.2&22.4 ms &1.4$\times$&78.4&279M \\
			AttentiveNAS-A2&77.5&22.5 ms & 1.3$\times$ &78.8&317M\\	
			\textbf{	\sysname @cpu-A3}  & \textbf{79.5} & \textbf{18.9 ms}& \textbf{2.6$\times$}&79.6& 981M \\
			\hline
			FBNetV2-L1 &75.8&25.0 ms &1.2$\times$ &77.2& 325M\\
			FBNetV3-A &78.2 &27.7 ms&1.3$\times$& 79.1 & 357M \\
			\textbf{	\sysname @cpu-A4}   &\textbf{80.0} & \textbf{24.4 ms} &\textbf{ 2.4$\times$}& 80.1& 1267M \\
			\toprule
			\multicolumn{6}{c}{ \textbf{(b) Results on the Google Pixel 4 with TFLite}}\\
			\multirow{2}{*}{Model}  &Acc\%& \multicolumn{2}{c|}{Pixel4 Latency}&Acc\%& \multirow{2}{*}{FLOPs} \\
			&\textbf{ INT8}& \textbf{INT8}&\textbf{speedup$^*$} &   FP32&\\ 
			
			\midrule
			
			MobileNetV3Small & 66.3&6.4 ms & 1.3$\times$& 67.4 &56M \\
			\textbf{	\sysname @pixel4-A0}& \textbf{73.6} & \textbf{5.9 ms} & \textbf{2.1$\times$}  &73.7 & 107M \\
			\hline
			MobileNetV2  &71.4&16.5 ms  & 1.9$\times$&72.0&300M\\
			ProxylessNAS-R  &74.6 &18.4 ms &1.8$\times$&74.6&320M  \\
			
			MobileNetV3Large   &74.5&15.7 ms &1.5$\times$&75.2&219M \\
			APQ-8bit & 74.6&14.9 ms &2.0$\times$ &74.4&340M\\
			OFA (\#25) &75.6&14.8 ms & 1.7$\times$&76.4 &230M\\
			OQAT-8bit & 75.8 & 15.2 ms & 1.9$\times$&76.2 &287M\\
			AttentiveNAS-A0 &76.1 &15.2 ms  &2.0$\times$& 77.3& 203M \\
			\textbf{	\sysname @pixel4-A1}& \textbf{77.6} & \textbf{14.7 ms} & \textbf{2.2$\times$} & 77.7& 274M  \\
			\hline
			APQ-8bit  &75.1 & 20.0 ms &1.9$\times$&75.1&398M \\
			OQAT-8bit  & 76.5 & 20.4 ms & 1.8$\times$&76.8&347M\\
			AttentiveNAS-A1&77.2&21.1 ms & 2.0$\times$&78.4&279M\\
			AttentiveNAS-A2&77.5&22.7 ms & 2.0$\times$ &78.8&317M\\	
			
			\textbf{	\sysname @pixel4-A2} & \textbf{78.3} & \textbf{19.4 ms} & \textbf{2.3$\times$} & 78.4& 402M\\
			\hline
			
			FBNetV2-L1 &75.8&26.7 ms &1.5$\times$&77.2& 325M \\
			OQAT-8bit  & 77.0 & 29.9 ms & 1.7$\times$&77.2&443M\\
			FBNetV3-A  &78.2 &30.5 ms & 1.5$\times$& 79.1& 357M\\
			\textbf{	\sysname @pixel4-A3}  & \textbf{79.5} & \textbf{30.8 ms} & \textbf{2.1$\times$} & 79.5& 591M\\
			\hline
			EfficientNet-B0 &76.7&36.4 ms &1.7$\times$&77.3&390M \\
			\textbf{	\sysname @pixel4-A4}   & \textbf{79.9} & \textbf{35.5 ms} & \textbf{2.2$\times$} &80.0& 738M\\				
			\hline
		\end{tabular}
		\label{tbl:endtoend}
	\end{table}

\noindent\textbf{Search cost}.  As depicted in Fig.~\ref{fig:searchcost}, our algorithm, {\algname}, is designed to be lightweight and suitable for real-world usage, requiring only 25 GPU hours to search a space of 5000 iterations. This remarkable speed is mainly due to our block-wise search space quantization scheme, which significantly reduces the cost of search space quality evaluation. In comparison, Fig.~\ref{fig:searchcost} demonstrates that training each search space from scratch without this scheme would consume an impractical 1200k GPU hours.


\subsection{The Effectiveness of  Discovered INT8 Models}
\label{sec:mainresult}
In this section, we demonstrate that our searched spaces deliver state-of-the-art quantized models. We compare with two strong baselines: (1) \textit{prior art manually-designed and NAS-searched models}; and (2) \textit{quantization-aware NAS}. For baseline (1), we collect official pre-trained FP32 checkpoints and conduct LSQ+ QAT to get the quantized accuracy. The hyperparameter settings follow the original LSQ+ paper, except that we set a larger epoch of 10 to achieve better accuracy. The latency numbers are measured on our devices. For (2), we compare with strong baselines including APQ~\cite{apq} and OQAT~\cite{oqa}. Specifically, we limit APQ to search for the fixed 8bit (INT8) models. Since OQAT has no 8bit supernet checkpoint, we follow the official source code and conduct supernet QAT for 50 epochs. The final INT8 models are searched under the same INT8 latency predictors for fair comparison.

\noindent\textbf{Results}.  Table~\ref{tbl:endtoend}  summarizes comparison results. Remarkably, our searched model family,  {\sysname} significantly outperform SOTA efficient models and quantization-aware NAS searched models, with higher INT8 quantized accuracy, lower INT8 latency and better speedups.
Without finetuning, our tiny models {\sysname}@cpu-A0 and  {\sysname}@pixel4-A0 achieve 74.7\% and 73.6\% top1 accuracy on ImageNet, which is 8.4\% and 7.3\% higher than MobileNetV3-Small (56M FLOPs) while maintaining the same level quantized latency. For larger models, {\sysname}@cpu-A4 (80.0\%) outperforms FBNetV3-A with 1.8\% higher accuracy while runs 3.3ms faster.  In particular, to achieve the same level accuracy (i.e., around 77.2\%), AttentiveNAS-A1 has 22.4ms latency while 
 {\sysname}@cpu-A1 (77.4\%) only needs 8.8 ms (2.6 $\times$ faster). More importantly, our searched models can better utilize the INT8 hardware optimizations:  the latency speedups compared to full-precision inference are all larger than 2$\times$, and this leaves room to search large-size models with higher accuracy.

\vspace{-1.5ex}
\subsection{Ablation Study}
\vspace{-1.5ex}

\begin{figure}[t]
	\centering
	\includegraphics[width=0.8\linewidth]{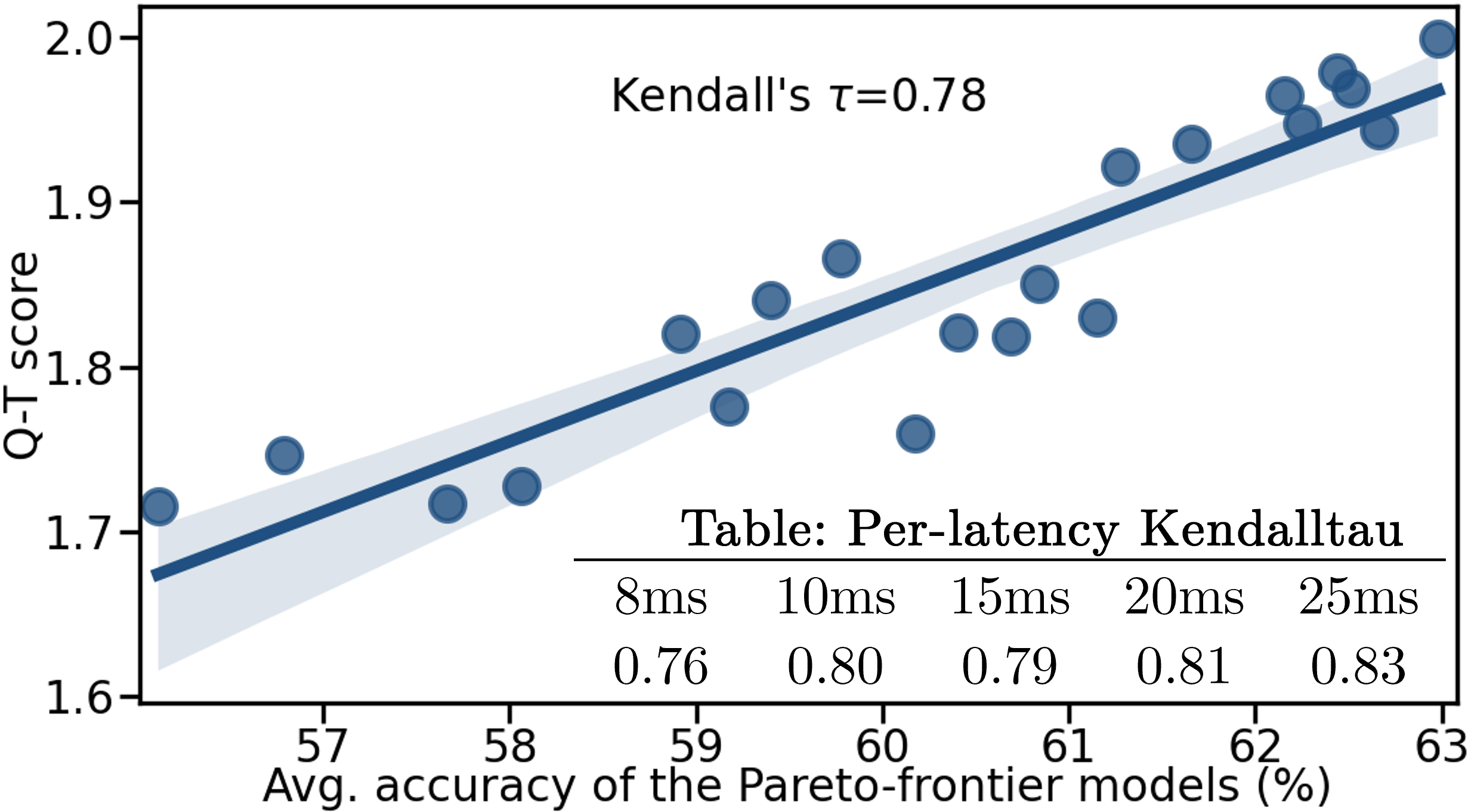}
	\vspace{-1.5ex}
	\caption{Q-T score  effectiveness (Kendall's $\tau$) on ranking search space quality. We achieve a high space ranking correlation. }
	\label{fig:rankingmodel}
\end{figure}

\noindent\textbf{Q-T  score effectiveness}. Q-T score is crucial as it guides the space evolution process. To evaluate its effectiveness, we randomly sample 30 search spaces, and measure the rank correlation (Kendall's $\tau$) between their Q-T score and their actual Pareto-frontier models' accuracies. Specifically,  we use Intel CPU as the test device and set a same latency constraints of \{8, 10, 15, 20, 25\}ms. For each sampled space, we train it from scratch for 50 epochs, and conduct evolutionary search to get the Pareto-frontier models' accuracies. As shown in Fig.~\ref{fig:rankingmodel}, the Kendall's $\tau$ between the Q-T score and the actual Pareto-frontier models' accuracies is 0.8, which indicates a very high rank correlation. 


\begin{table}[t]
	\centering
	\fontsize{8.5}{8.5} \selectfont
	\caption{Different space search methods and their best resulting quantized models on Pixel4. Baseline is a SOTA mobile-friendly AttentiveNAS space. $^*$: the search dimension use the same settings in AttentiveNAS. }
	\vspace{-2.5ex}
	\begin{tabular}
		{@{\hspace{0.2\tabcolsep}}c@{\hspace{0.8\tabcolsep}}c@{\hspace{1\tabcolsep}}c@{\hspace{1\tabcolsep}}c@{\hspace{1\tabcolsep}}c@{\hspace{1\tabcolsep}}c@{\hspace{1\tabcolsep}}c@{\hspace{1\tabcolsep}}c@{\hspace{0.2\tabcolsep}}}
		\hline
		\multirow{2}{*}{Method} &\multirow{2}{*}{Op} & \multirow{2}{*}{Width} &\multicolumn{5}{c}{Best quantized models}\\
		& & &10ms& 15ms & 20ms  &30ms & 36ms\\
		
		\hline 
		Baseline&-&-&-&76.6&77.2 &79.0&79.5\\
		{\algname}-op & search & fix$^*$&75.0&76.6&77.8 &78.6&78.8\\
		{\algname}-width &fix$^*$ &search& 75.4&77.4&78.0  &79.1&79.5\\
		{\algname} &search&search&\textbf{75.7}&\textbf{77.6} & \textbf{78.3} &\textbf{79.5}&\textbf{79.9}\\
		\hline
	\end{tabular}
	\label{tbl:search_compare}
\end{table}

\noindent\textbf{Ablation study on two search dimensions}. In Sec.~\ref{sec:analysis}, we conclude that operator type and configuration are two key factors impacting INT8 latency, which serves as the two search objectives of {\algname}. To verify the effectiveness, we create two strong baselines based on  the SOTA edge-regime AttentiveNAS search space: (1) {\algname}-op: we fix each elastic stage's width to AttentiveNAS  space, then allow each elastic stage to search for the optimal operator; and (2) {\algname}-width: we fix all elastic stages' block types to AttentiveNAS space, then  search for the optimal width. Table~\ref{tbl:search_compare} reports the space comparison between different search methods on the Pixel4. By searching both operator type and width, {\algname} finds the optimal search space where its best searched quantized models achieve the highest accuracy under all latency constraints. Moreover, even searching for one dimension, {\algname}-op and {\algname}-width outperform  the manually-designed AttentiveNAS space under small latency constraints.

\noindent\textbf{Search space design implications}. We now summarize our
learned experience and implications for designing quantization-friendly search spaces. We notice that the searched spaces show different preferences when targeting different devices: (i) All stages should use much wider channel widths compared to existing manually-designed spaces on the cpu, while only early stages prefer wider channels on Pixel  4. (ii) Since SE and Swish are INT8 latency-friendly on mobile phones, so our auto-generated search spaces for Pixel4 have many MBv3 stages. On Intel CPU, INT8 quantization slows down SE, Hardwish, and Swish, making FusedMB and MBv2 the priority for search spaces, with only the last two stages using MBv3. The  details are provided in supplementary.

\vspace{-1ex}
\section{Conclusion}
\vspace{-1ex}
In this paper, we introduced {\algname}, the first to automatically design a quantization-friendly  space for target device, which delivers superior INT8 quantized models with SOTA efficiency on real-world edge devices. 
Extensive experiments on two popular devices demonstrate its effectiveness compared to prior art manual-designed search spaces. 
We plan to apply {\algname} to other hardware efficiency such as energy-efficient search space design in the future.

\balance
{\small
	\bibliographystyle{ieee_fullname}
	\bibliography{ref}
}

\end{document}